\documentclass[mlmain]{jmlr}

\usepackage{longtable}
\usepackage{booktabs}
\usepackage[load-configurations=version-1]{siunitx}

\usepackage{tikz}
\usetikzlibrary{arrows.meta,positioning,calc,fit,shapes.multipart, shapes.geometric,backgrounds,calc}

\usepackage{parskip}

\theorembodyfont{\upshape}
\theoremheaderfont{\scshape}
\theorempostheader{:}
\theoremsep{\newline}

\jmlrvolume{}
\firstpageno{1}

\jmlryear{2025}
\jmlrworkshop{Symmetry and Geometry in Neural Representations}

\title[Lie‑Pseudogroup Structure in Algorithmic Agents]{%
Compositional Symmetry as Compression:\\
Lie‑Pseudogroup Structure in Algorithmic Agents
}

   \author{\Name{Giulio Ruffini} \Email{giulio.ruffini@neuroelectrics.com}
   \\
   \addr Neuroelectrics, Starlab, BCOM (Barcelona, Spain)}

\begin{document}

\maketitle

\begin{abstract}
In the algorithmic (Kolmogorov) view, agents are programs that track and compress sensory streams using  generative programs. We propose a framework where the relevant structural prior is simplicity (Solomonoff) as \emph{compositional symmetry}, where natural streams are well described by (local) actions of finite‑parameter Lie pseudogroups on geometrically and topologically complex low‑dimensional configuration manifolds (latent spaces). Modeling the agent as a generic neural dynamical system coupled to such streams, we show that accurate world‑tracking imposes (i) \emph{structural} constraints (equivariance of the agent system constitutive equations and readouts) and (ii) \emph{dynamical} constraints: under static inputs, symmetry induces conserved quantities (Noether‑style labels) in agent dynamics and confines trajectories to reduced invariant manifolds; under slow drift, these manifolds move but remain low‑dimensional. This yields a hierarchy of reduced manifolds aligned with the compositional factorization of the pseudogroup---a geometric account of the ``blessing of compositionality'' in deep models. We connect these ideas, at a high level, to the Spencer formalism for Lie pseudogroups, and formulate a symmetry‑based, self‑contained version of predictive coding in which higher layers receive only \emph{coarse-grained residual transformations} (prediction‑error coordinates) along symmetry directions unresolved at lower layers.
\end{abstract}

\begin{keywords}
Kolmogorov Theory of Consciousness, Lie pseudogroups, Compositional symmetry, Kolmogorov complexity, Hierarchical reduction, manifold hypothesis, coarse-graining, predictive coding, compression
\end{keywords}

\section{Introduction}
\label{sec:intro}
\begin{figure} [t]
    \centering
    \includegraphics[width=0.71\linewidth]{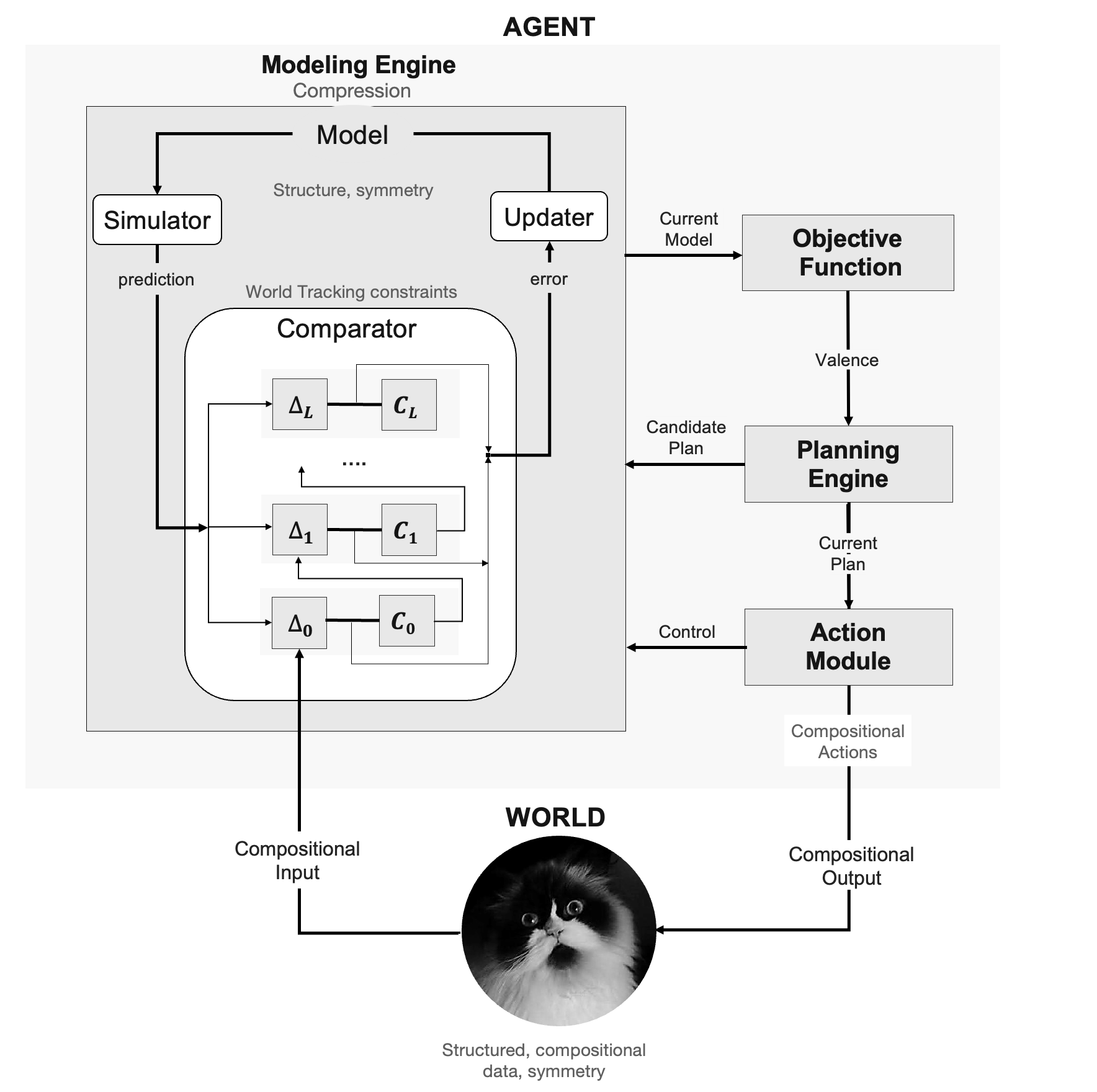}

\caption{
The \textbf{algorithmic agent}  interacts with the World (structure, symmetry, compositional data) \citep{ruffiniAlgorithmicInformationTheory2017, ruffiniAITFoundationsStructured2022, ruffiniAlgorithmicAgentPerspective2024}. The~\textit{Modeling Engine} (compression) runs the current Model (which encodes found structure/symmetry), makes hierarchical/compositional predictions of future coarse-grained (pooled) data, and then evaluates the prediction error in the \textit{Comparator} (world-tracking constraint monitoring) to hierarchically update the Model. The comparison process is carried \textit{hierarchically}, and the output is coarse-grained to feed the next level.   We reflect this process mathematically as a world-tracking constraint on the dynamics (Equation~(\ref{eq:track})
\textbf{Hierarchical modeling engine (Comparator).} Level $k$ predicts $\hat I_k=\hat\gamma_k\!\cdot I_0$, compares to its incoming datum (raw image at $k=0$; canonicalized, coarse‑grained residual $m_{k-1\to k}$ for $k\ge1$), updates from error $E_k$, and forwards only the residual after canonicalization and coarse‑graining, $m_{k\to k+1}=\mathcal{C}_{k\to k+1}(\hat\gamma_k^{-1}\!\cdot\hspace{0.12em}\text{input}-I_0)$. Generators shrink along $H_0\supset H_1\supset H_2 \cdots$; residuals live in quotient directions and carry “what’s left to explain” to coarser scales. The~\textit{Planning Engine} runs counterfactual simulations and selects plans for the next (compositional) actions (agent outputs to world and self). The~Updater receives hierarchical prediction errors from the Comparator as inputs to improve the Model. All modules can be implemented hierarchically. }
\label{fig:hier_model_engine}
\end{figure}

Kolmogorov Theory (KT) proposes a scientific framework centered on algorithmic agents using the mathematics of algorithmic information theory (AIT), where the central concepts are computation and compression. In particular, the Kolmogorov complexity of a data object--- the length of the shortest program producing the data object \citep{coverElementsInformationTheory2006}---provides a powerful conceptual anchor in the theory.
In KT, an algorithmic agent is an information‑processing system that builds and runs \emph{compressive} models (programs) of its environment, using them to guide action and shape experience  (see Figure~\ref{fig:hier_model_engine}). Model building serves an Objective Function (telehomeostasis, or preservation of pattern, in living systems) and is guided by Ockham's Razor (simpler models/shorter programs are preferred)~\citep{soberOckhamsRazorsUsers2015, ruffiniAITFoundationsStructured2022,ruffiniStructuredDynamicsAlgorithmic2025}. 

A \textit{model} is the \emph{short program}  that generates a given dataset. The same program can be read as a smooth generative map \(f:\mathcal C\to\mathbb R^X\), turning parameters into data object instances. A third view is symmetry‑based: the generator induces a group of automorphisms whose orbits cover the data manifold; learning invariances recovers the generator~\citep{ruffiniModelsNetworksAlgorithmic2016}. When no short deterministic program exists, the optimal compressor is statistical; expected Kolmogorov complexity converges to entropy rate~\citep{coverElementsInformationTheory2006,liApplicationsAlgorithmicInformation2007}.

A central question is: \emph{what structure can such compressive models exploit?} 
The language of group theory provides a good handle to formalize the notion of structure. Discrete groups admit \emph{Cayley graphs} encoding generators and relations~\citep{biggs74}. For Lie groups, infinitesimal structure lies in a finite‑dimensional Lie algebra; for semisimple algebras, \emph{Dynkin diagrams} encode simple roots and relations~\citep{humphreys72,knapp02}. Lie \emph{pseudogroups}---local transformation groups defined by involutive PDEs---admit infinitesimal descriptions that can be organized in the \emph{Spencer complex}, whose differentials encode compatibility conditions~\citep{goldschmidt67,seiler10}.

Natural streams exhibit \emph{compositional symmetry}: pose, viewpoint, articulations, and semantic deformations compose recursively and act locally \citep{simonArchitectureComplexity1991,riesenhuberHierarchicalModelsObject1999,poggioTheoreticalIssuesDeep2020,cagnettaHowDeepNeural2024,ruffiniStructuredDynamicsAlgorithmic2025}. This points beyond global (finite) Lie groups to \emph{Lie pseudogroups} of local diffeomorphisms as a language to describe structure, which can be captured (when needed) in the Spencer framework.
This framework provides a jet‑level \emph{compatibility complex} (the Spencer complex)—a quiver‑like graded sequence of bundles and differential operators whose cohomology governs formal integrability, prolongation, and obstructions—organising how local symmetries compose and constrain dynamics across scales.
In plainer terms, it acts as a structured \emph{consistency checker}: it tracks, order by order in derivatives, whether local symmetry rules can be stitched into a coherent global transformation, and pinpoints exactly where that stitching fails. 

\begin{figure}[t]
\centering
\resizebox{\textwidth}{!}{%
\begin{tikzpicture}[
  x=1cm, y=1cm, transform shape,
  font=\Large,
  >=Latex,
  box/.style={rounded corners=3pt, draw=black, very thick, align=left,
              fill=white, inner sep=6pt, minimum width=8.2cm, minimum height=2.8cm},
  thinbox/.style={rounded corners=3pt, draw=black, thick, align=left,
                  fill=white, inner sep=6pt, minimum width=8.2cm, minimum height=2.8cm},
  arrow/.style={-Latex, ultra thick},
  dasharrow/.style={-Latex, thick, dashed},
  lbl/.style={font=\normalsize, inner sep=1pt, fill=white},
  rowband/.style={fill=black!5, draw=none, rounded corners=4pt}
]

\def\xA{0}
\def\xB{10}
\def\xC{20}
\def\yTop{0}
\def\yMid{-8}
\def\yBot{-16}

\node[box] (gen) at (\xA,\yTop) {%
\textbf{Generative model} (\S\ref{sec:lie_generative})\\[2pt]
$f:\mathcal C\to\mathbb R^{X},\quad I=f(c)$\\
$c=\gamma\!\cdot c_0,\quad I=\gamma\!\cdot I_0,\ \ \gamma\in G$
};

\node[box] (lie) at (\xB,\yTop) {%
\textbf{Lie pseudogroup $G$}\\[2pt]
$\displaystyle \gamma=\exp\Bigl(\sum_k \theta_k T^k\Bigr)$\\
Compositional symmetry (local)
};

\node[thinbox] (spencer) at (\xC,\yTop) {%
\textbf{Spencer/compatibility}\\[2pt]
Hierarchy of constraints\\
(integrability, exactness)
};

\draw[arrow] (lie) -- node[lbl, above] {acts on $\mathcal C,\ \mathbb R^{X}$} (gen);
\draw[arrow] (lie) -- (spencer);

\node[box] (wt) at (\xA,\yMid) {%
\textbf{World--tracking dynamics} (\S\ref{sec:world_tracking})\\[2pt]
$\dot x=f(x;w,I_{\theta(t)}),\qquad p(x)\approx I_{\theta(t)}$};

\node[box] (equiv) at (\xB,\yMid+3) {%
\textbf{Equivariance}\\[2pt]
$f(\gamma\!\cdot x;w,\gamma\!\cdot I)=\gamma\!\cdot f(x;w,I)$\\
$p(\gamma\!\cdot x)=\gamma\!\cdot p(x)$\vspace{1pt}
};

\node[box] (noether) at (\xC-5,\yMid-1) {%
\textbf{Noether--style invariants}\\[2pt]
Static $I_{\theta_\star}$: $\displaystyle \frac{d}{dt}\,p(x)=0$\\
Conserved labels $\Rightarrow$ reduced leaf
};

\draw[arrow] (gen) -- (wt);
\draw[arrow] (lie) -- (equiv);
\draw[arrow] (wt) -- (noether);
\draw[arrow] (equiv) -- (noether);

\draw[dasharrow] (equiv.west) to[bend right=20] node[lbl, above left] {structural constraints on $w$} (wt.east);

\node[box] (hier) at (\xA,\yBot+2) {%
\textbf{Hierarchical coarse--graining} (\S\ref{ssec:lie_hierarchy})\\[2pt]
$G=H_0\supset H_1\supset\cdots\supset H_L,\qquad
\mathcal M_k=\mathcal M_{k-1}/H_k$\\[2pt]
\emph{Canonicalize \& coarse--grain residual:}\\
$r_k=\hat\gamma_k^{-1}\!\cdot I - I_0,\quad
m_{k\!\to\!k+1}=\mathcal C_{k\!\to\!k+1}(r_k)$
};

\node[box] (resid) at (\xB-7,\yBot-3) {%
\textbf{Predictive residuals} (\S\ref{ssec:predictive_lie})\\[2pt]
$\varepsilon_k=\hat\gamma_k^{-1}\gamma_{k-1}\in H_{k-1}/H_k,\quad
\varepsilon_k=\exp(\eta_k^a S_k^a)$\\[2pt]
\emph{Project \& pass up:}\;\; $\eta_k=P_k\eta_k+Q_k\eta_k$;\; send $Q_k\eta_k$\\
(Message $m_{k\!\to\!k+1}$ encodes quotient directions $H_{k-1}/H_k$)
};

\node[box] (reduced) at (\xC,\yBot-3) {%
\textbf{Compositional reduced dynamics}\\[2pt]
Nested invariant manifolds $\{\mathcal M_k\}$\\
“Blessing of compositionality”
};

\draw[arrow] (noether) -- (hier);
\draw[arrow] (spencer) |- (hier);

\draw[arrow] (hier) to[bend left=12]
     node[lbl, above] {quotients define $\{\mathcal M_k\}$} (reduced);

\draw[dasharrow] (resid) -- 
     node[lbl, above] {drift/updates on $\mathcal M_k$} (reduced);

\draw[arrow] (hier) -- node[lbl, above] {message $m_{k\!\to\!k+1}$} (resid);

\begin{pgfonlayer}{background}
  \node[rowband, fit=(gen) (lie) (spencer), inner sep=10pt] {};
  \node[rowband, fit=(wt) (equiv) (noether), inner sep=10pt] {};
  \node[rowband, fit=(hier) (resid) (reduced), inner sep=10pt] {};
\end{pgfonlayer}

\end{tikzpicture}%
} 
\caption{\textbf{From generative symmetry to compositional reduced dynamics.} Top: a Lie pseudogroup \(G\) acts locally on configuration \(\mathcal C\) and observation space, with Spencer providing a hierarchy of compatibility constraints. Middle: the agent enforces world tracking (Eq.~\eqref{eq:track}); equivariance (Eq.~\eqref{eq:equiv}) imposes structural constraints and, for static inputs, yields conserved labels (Noether‑style), confining trajectories to reduced leaves. Bottom: a flag \(G=H_0\supset\cdots\supset H_L\) \emph{defines} the nested invariant manifolds \(\{\mathcal M_k\}\) (solid bent arrow); residual messages \(m_{k\to k+1}\) parameterize quotient directions \(\varepsilon_k\in H_{k-1}/H_k\) and \emph{induce drift/updates} on these leaves (dashed arrow), yielding a hierarchy of compositional reduced dynamical manifolds.}
\label{fig:lie_pipeline_fullpage}
\end{figure}
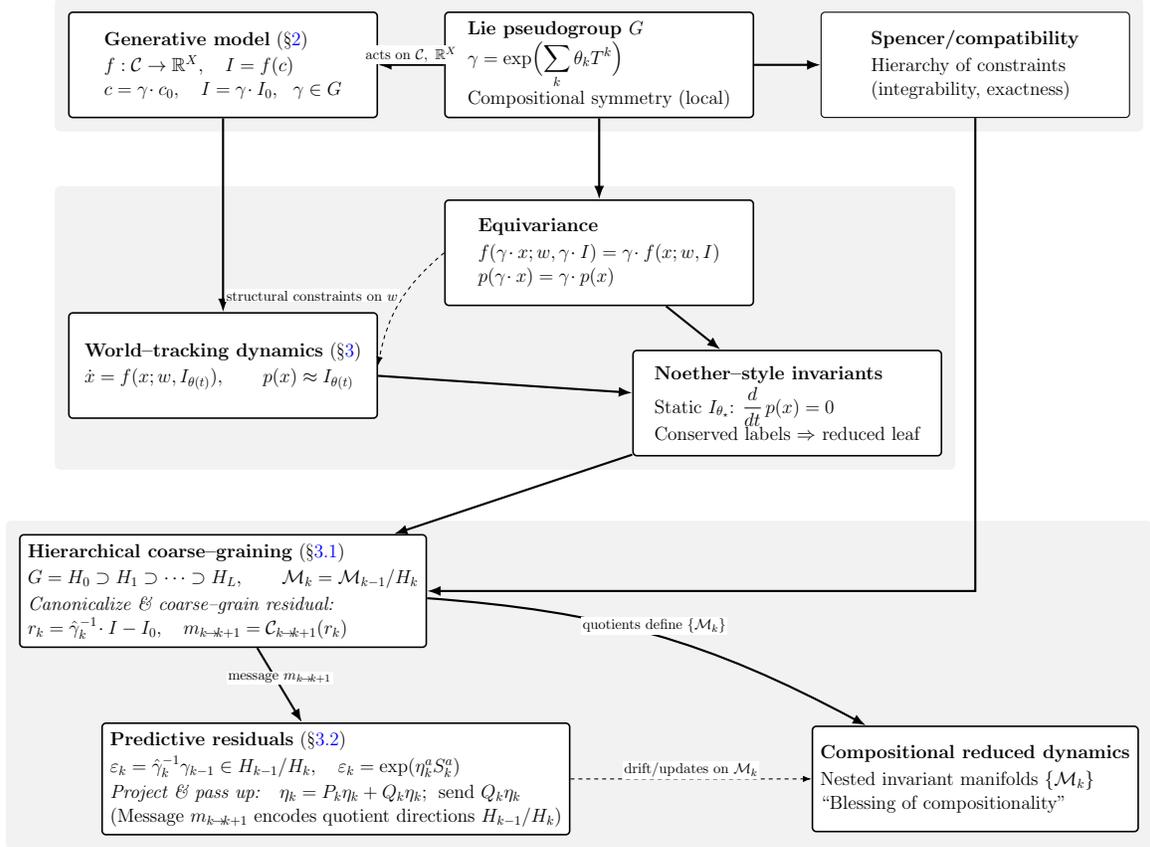

We  take the persepective of Lie pseudogroups as a programming language  to describe generative models and to motivate a built‑in  modeling hierarchy.
Here we propose a symmetry‑first account of model structure for algorithmic agents where we (i) define \emph{generative models} as local actions of finite‑parameter Lie pseudogroups on configuration manifolds, (ii) model the agent as a neural ODE driven by such streams with a \textit{Comparator} enforcing tracking, and  (iii) derive the \emph{constitutive} (equivariance) and \emph{dynamical} (Noether‑style invariants, reduced manifolds) constraints that follow. This structure explains why deep, hierarchical architectures---which mirror compositional symmetry---attain favorable sample complexity on hierarchical tasks~\citep{poggioWhyWhenCan2016,poggioCompositionalSparsityLearnable2024}, and why a \emph{bare} manifold prior can be insufficient without additional geometric covering structure~\citep{kiani2024hardness}. The analysis suggests symmetry‑aware designs (e.g., group‑equivariant networks) as principled architectures for agents and offers a group-theoretical lens on predictive coding.
Finally, we provide a tentative implementation of hierarchical predictive processing \cite{fristonDoesPredictiveCoding2018}  using this formalism. 
For an overview of the logical structure of the paper, see Figure~\ref{fig:lie_pipeline_fullpage}.

\paragraph{Intuition.}
Compositional symmetry is a \emph{program schema}: complex transformations are built by nesting a few primitive moves near the identity. Agents that align their internal structure with this recursive grammar compress better, learn with fewer samples, and generalize over orbits of the same symmetry.

\section{Generative Models as Lie‑Pseudogroup Actions}
\label{sec:lie_generative}

Natural data will appear low‑dimensional (the manifold hypothesis) if generated by \emph{hierarchical continuous symmetries} generated by a finite dimensional Lie pseudogroup. We capture this through the local actions of a finite‑parameter Lie pseudogroup \(G\) on a configuration manifold \(\mathcal C\):

\begin{definition}[Generative model]
A \emph{generative model} is a smooth map \(f:\mathcal C\to\mathbb R^X\) from an \(M\)-dimensional configuration manifold (\(M\ll X\)) to observations. For \(c\in\mathcal C\), write \(I=f(c)\in\mathbb R^X\).
\end{definition}

Let \(G\) act locally on \(\mathcal C\) and on \(\mathbb R^X\). Pick \(c_0\in\mathcal C\) with \(I_0=f(c_0)\).

\begin{definition}[Lie generative model]
\label{def:lie_gen_model}
The map \(f\) (or \(f(\mathcal C)\)) is a \emph{Lie generative model} if
\[
\forall c\in\mathcal C\;\exists\,\gamma\in G:\quad c=\gamma\!\cdot\!c_0,\quad f(c)=\gamma\!\cdot\!I_0.
\]
Equivalently, \(I=f(\gamma\!\cdot\!c_0)=\gamma\!\cdot I_0\) with \(\gamma\in G\).
\end{definition}

\paragraph{Intuition (recursion and compression).}
Because \(G\) is a Lie (pseudo)group, elements near the identity factor as
\[
\gamma=\exp\Bigl(\sum_{k=1}^r\theta_k T^k\Bigr),
\]
so complex deformations arise by composing \emph{infinitesimal} ones~\citep{hall2015}.
Here $\gamma(\cdot):U\to G$ is a local parameterization of $G$, and “$\cdot$” denotes the induced (generally nonlinear) action of $G$ on image space.
This endows \(f\) with a recursive, compositional parameterization: a short, structured program in the sense of algorithmic information theory. Pseudogroups permit this \emph{locally} even when global topology precludes a single global action~\citep{seiler10}.

If natural data is generated by a ``random walk" through the parameter space of a Lie pseudogroup, the following apply: (i) \emph{Compression:} the data can be compressed, because \(r\) parameters span the data manifold. (ii) \emph{Efficient learning:} deep architectures that respect generators can enjoy the \textit{blessing of compositionality}~\citep{poggioWhyWhenCan2016,poggioCompositionalSparsityLearnable2024,miaoLearningLieGroups2007,anselmiRepresentationLearningSensory2022}---fewer data (sample complexity) is needed for learning with the right architecture). (iii) \emph{Symmetry discovery:} when \(G\) is unknown, its generators can be potentially be inferred~\citep{Moskalev2022-qf}. We study the dynamical implications for the world-tracking agent in the next section.

\section{Symmetry-Constrained World-Tracking Dynamics}
\label{sec:world_tracking}

Here, we assume natural input/sensory streams are not arbitrary but arise from finite-parameter Lie
(pseudo)group generators acting on a low-dimensional configuration manifold
(Sec.~\ref{sec:lie_generative}) and show that an
algorithmic agent that can \emph{track} such
streams need to be compatible with their underlying symmetry.   
Let the external data stream be generated by a finite‑parameter Lie pseudogroup $G$
acting on a reference image $I_{0}$,
\begin{equation}\label{eq:world}
  I_{\theta(t)} \;=\; \gamma(\theta(t))\!\cdot\! I_{0}, 
  \qquad \theta(t)\in U\subset\mathbb R^{r}\ \text{(local chart on $G$)}.
\end{equation}

The agent’s high‑dimensional state $x\in\mathbb R^{X}$ follows the general neural network equation
\begin{equation}\label{eq:fast}
  \dot x = F\!\bigl(x;\,w,\,I_{\theta(t)}\bigr),
\end{equation}
with fixed weights $w$.  
A projector $p:\mathbb R^{X}\!\to\!\mathbb R^{Y}$ ($Y\!\ll\!X$) provides the
\emph{world-tracking equation}
\begin{equation}\label{eq:track}
  p\bigl(x(t)\bigr)\;\approx\;I_{\theta(t)} .
\end{equation}
This equation states that the agent is able to lock into the input data stream.

\paragraph{Compatibility of tracking with invariants requires equivariance.}
Equivariance of Equation~\ref{eq:fast} says “solutions come in \(G\)–families.” Uniqueness lets us \emph{name} each solution by the group element needed to reach it from a reference (for simplicity we consider connected groups). These names are constant along trajectories (conserved). 

The tracking constraint (Equation~\ref{eq:track}) is feasible precisely when it latches onto these constants; otherwise it fights the flow.
Under static inputs the constraint \(p(x)=I_{\theta_\star}\) can be sustained only if it \emph{depends on} the conserved labels (cyclic coordinates) induced by the symmetry of the equations. Otherwise it would over‑constrain trajectories to a point. In canonical coordinates adapted to the symmetry, \(p(x)\) must be a function of the corresponding constants of motion. With slowly varying \(\theta(t)\) the labels become \emph{adiabatic invariants}, and the constrained leaf \emph{drifts} but remains low‑dimensional.

\vspace{-3pt}
\paragraph{Consequences of equivariance requirement.}
Because $I_{\theta}$ is moved by $G$ (the equation must hold for any $\gamma$), effective tracking demands that the
internal dynamics respect the same action,
\begin{equation}\label{eq:equiv}
  \forall\gamma\in G:\;
  f\!\bigl(\gamma\!\cdot\!x;\,w,\gamma\!\cdot\!I_\theta\bigr)
  = \gamma\!\cdot\!f\!\bigl(x;\,w,I_\theta\bigr),\quad
  p(\gamma\!\cdot\!x)=\gamma\!\cdot\!p(x).
\end{equation}
Eq.~\eqref{eq:equiv} places \emph{structural constraints} on $w$ (weight
tying, zero blocks, etc.) identical in spirit to
group‑equivariant CNNs \citep{Moskalev2022-qf,ruffiniStructuredDynamicsAlgorithmic2025}.
For infinitesimal $\gamma = \exp(\epsilon T_{k})$ we obtain linear
commutation conditions
$
  T_{k}f = (\partial f/\partial x)T_{k}x
$
that must hold for every Lie‑algebra generator $T_{k}$.

\vspace{-3pt}
\paragraph{Conservation laws.}
Freeze $\theta$ so that $\gamma(t)=\gamma_{0}$.  
Under Eq.~\eqref{eq:equiv} the read‑out becomes invariant:
$p(x)=\text{const}$.
Hence, each of the $Y$ read‑out channels defines a conserved quantity.
This is a direct Noether analogue: continuous symmetry
$\gamma\!\cdot$   $\Rightarrow$ invariants
$\bigl\{p_{j}(x)\bigr\}_{j=1}^{Y}$
 \citep{Hydon2000-kp}.
Trajectories are confined to an $(X{-}Y)$‑dimensional leaf of phase
space.
When $\theta(t)$ varies slowly, the invariant leaf drifts, but its dimension
remains $\le M+X-Y$ (gains at most $M$ extra degrees of freedom).

\vspace{-3pt}
\paragraph{Approximate tracking via Lyapunov control.} To formalize the notion of approximate tracking after transient dynamics ($p\bigl(x\bigr)\;\approx\;I_{\theta} $), we can use the machinery of Lyapunov functions. 
We modify the constrained dynamical equations above by defining error and Lyapunov functions and add a symmetry‑preserving feedback term
\begin{equation} \label{eq:feedback}
  E = p(x)-I_{\theta}, \quad
  V=\tfrac12\|E\|^{2}, \quad  \dot x = F\bigl(x;w,I_{\theta}\bigr) + K\,E,
\end{equation}
with a gain operator $K$ that commutes with every $T_{k}$.
We then need to choose a gain $K$ such that $D_x p(x)\,K$ is negative semi-definite and $K$ commutes with the induced action of each generator $T_k$ (so the tracking equations remain equivariant). Then $\dot V \le 0$ under static inputs; with slow drift one obtains an ISS-type bound $\dot V \le -\alpha\|E\|^2 + \beta\|\dot I_\theta\|^2$ for suitable $\alpha,\beta>0$ \citep{sontagInputStateStability2008}.

\noindent\textbf{Remark.} All constructions are local: actions, exponentials, and invariant leaves are taken in the chart domains where the pseudogroup is defined; global statements require additional compatibility (Spencer exactness).
  
\vspace{-3pt}
\paragraph{Summary.}
Equivariance (\ref{eq:equiv}) forces \textbf{structural} constraints on the
agent (weights obey group commutation) and \textbf{dynamical} constraints
(conserved read‑outs and low‑dimensional invariant manifolds).  A Lyapunov
closure (\ref{eq:feedback}) formalises the “$\approx$’’ in
(\ref{eq:track}), completing a compact, symmetry‑first description of
world‑tracking dynamics.

\subsection{Lie‑pseudogroups, hierarchical coarse‑graining, and reduced manifolds}
\label{ssec:lie_hierarchy}

Compositionality enters via generator nesting: every \(\gamma\in G\) near the identity factors as \(\exp(\sum_k\theta_k T^k)\). A \emph{hierarchical description} is obtained by declaring certain generators negligible at coarser levels. Formally, choose a flag of sub‑pseudogroups
\begin{equation}
G=H_0\supset H_1\supset\cdots\supset H_L,
\end{equation}
where \(H_{k}\) is obtained from \(H_{k-1}\) by omitting some generators. Let \(\mathcal M_0\) denote the full state space; define the reduced manifold at level \(k\) as the quotient \(\mathcal M_{k}:=\mathcal M_{k-1}/H_k\) (locally: flows of the discarded generators are frozen). Then
\begin{equation}
\mathcal M_0 \supset \mathcal M_1 \supset \cdots \supset \mathcal M_L,
\qquad
\dim \mathcal M_{k+1} < \dim \mathcal M_k,
\end{equation}
and motion within \(\mathcal M_k\) is driven only by residual generators in \(H_k\).
In the formal theory of differential constraints, the flag \(H_{0}\supset\dots\supset H_{L}\) is encoded by the Spencer sequence of \(G\); exactness of that sequence guarantees compatibility of the layered constraints and integrability of the reduced dynamics. Hence
{\small
\[ \text{Lie pseudogroup} \;\Longrightarrow\; \text{Spencer tower of sub‑pseudogroups} \;\Longrightarrow\; \text{nested invariant manifolds} \] 
}
Each quotient sharpens the agent’s description while retaining the ability to \emph{recompose} fine structure by re‑activating generators, capturing the “blessing of compositionality’’: models learn efficiently on \(\mathcal M_{k}\) yet can recover detail by ascending the tower. In this sense, hierarchical cognition, coarse–grained abstraction, and the algebraic–geometric machinery of Lie pseudogroups are three facets of the same structure.

\paragraph{Intuition.}
Each level ``throws out’’ a set of generators, fixes the associated conserved labels, and descends to a simpler manifold. The tower mirrors the grammar of \(G\): coarse levels use fewer letters; fine levels re‑introduce them as needed. Spencer’s framework encodes such hierarchical compatibility; exactness corresponds to the absence of obstructions in the layered constraints~\citep{seiler10}.

\subsection{Predictive hierarchy as residual transformations}
\label{ssec:predictive_lie}

Here, we provide a tentative formalization of the predictive processing hierarchy. 
As before, all statements are local in the domain where the pseudogroup action and charts overlap; residuals and compositions are taken in that neighborhood.

Incoming data are processed bottom-up (see Figure~\ref{fig:hier_model_engine}). The most detailed layer predicts fine features using its own generators, compares prediction to input, and forms an \emph{error}: the part it could not explain. That error is used locally to refine the layer’s hypothesis; only the \emph{unexplained remainder} is coarse-grained (pooled) and passed upward. The next layer absorbs what it can with its own generators and again forwards just the coarse-grained residual:
{\small
\[
\text{predict} \ \rightarrow\ \text{compare} \ \rightarrow\ \text{update locally} \ \rightarrow\ \text{canonicalize residual} \ \rightarrow\ \text{coarse\mbox{-}grain} \ \rightarrow\ \text{pass up}
\]
}
\noindent In steady scenes the residuals shrink toward zero at all levels, leaving stable labels (conserved coordinates); a persistent residual at the top flags a missing generator or a model mismatch.

\paragraph{Set--up and linear residual fit.}
Let $G=H_0\supset H_1\supset\cdots\supset H_L$ be a nested family of (local) sub\mbox{-}pseudogroups with Lie algebras $\mathfrak g\supset\mathfrak h_1\supset\cdots\supset\mathfrak h_L$. At time $t$ the datum is $I_{\theta(t)}=\gamma_{\mathrm{true}}(t)\!\cdot I_0$, while level~$k$ predicts $\hat I_k(t)=\hat\gamma_k(t)\!\cdot I_0$ with $\hat\gamma_k(t)\in H_k$. The observation--space error is
$
\delta I_k \;=\; I_{\theta(t)} - \hat I_k(t).
$
Because the stream is (locally) generated by a group action, small errors are explained by a small \emph{residual} $\varepsilon_k=\exp(\eta_k)$ near the identity, so that $I_{\theta(t)} \approx (\hat\gamma_k\,\varepsilon_k)\!\cdot I_0$. Linearizing the action at $\hat I_k$ with a basis $\{S_a\}$ of $\mathfrak g$ and induced image--space velocities $V_a(\hat I_k)$,
$
\delta I_k \;\approx\; \sum_a \eta_k^{\,a}\,V_a(\hat I_k),
$
so we obtain $\eta_k=(\eta_k^{\,a})$ by a regularized least--squares fit (e.g.,  Tikhonov regularization).

\paragraph{Within--level update via a symmetry--respecting projection.}
Level $k$ can realize only directions in $\mathfrak h_k=\mathrm{Lie}(H_k)$. Fix an inner product on $\mathfrak g$ that is $\mathrm{Ad}(H_k)$--invariant and let $P_k:\mathfrak g\to\mathfrak h_k$ be the associated orthogonal projector (with $Q_k:=I-P_k$). The realizable part updates the hypothesis, 
$
\hat\gamma_k \;\leftarrow\; \hat\gamma_k\,\exp\!\big(P_k\,\eta_k\big).
$
The \emph{unresolved} component $Q_k\,\eta_k$ is what must be communicated upward.

\paragraph{Canonicalize and coarse--grain before passing up.}
To present a clean message at the next scale, first express the residual in a common reference frame (canonicalization),
\begin{equation}
r_k \;:=\; \hat\gamma_k^{-1}\!\cdot I_{\theta(t)} \;-\; I_0,
\end{equation}
which removes the frame explained at level $k$. Then apply an $H_k$--invariant, $H_{k+1}$--equivariant coarse--grainer
$\mathcal C_{k\to k+1}:\mathbb R^{Y}\!\to\!\mathbb R^{Y_{k+1}}$ and define the upward message
\begin{equation}
m_{k\to k+1} \;:=\; \mathcal C_{k\to k+1}\!\big(r_k\big).
\end{equation}
Intuitively, $\mathcal C_{k\to k+1}$ “irons out’’ fine variation along $H_k$--orbits while preserving precisely the structure modelled at level $k{+}1$. The next layer treats $m_{k\to k+1}$ as its comparator datum and repeats the same loop with its generators. Equivalently, one may pass an algebra--space message $\bar\eta_{k\to k+1}=\mathrm{Ad}_{\hat\gamma_k^{-1}}(Q_k\,\eta_k)$; we use the observation--space version for numerical stability and alignment with predictive coding practice.

\section{Discussion}
\label{sec:discussion}

Starting from the premise that sensory streams are generated by finite‑parameter Lie (pseudo)group actions, we offer a symmetry‑aware, group-theoretic account of compression and tracking, with the additional benefit of mathematica tools to describe structure in models. Under static inputs, equivariance of the closed‑loop vector field yields Noether‑style conserved labels (solution‑symmetry invariants), and the tracking constraint is feasible precisely when it depends on those labels; trajectories lie on reduced invariant manifolds. Under slow drift, the labels become adiabatic, and the manifold drifts with at most \(M\) added degrees of freedom. Structural equivariance constrains parameters, guiding architecture design (e.g., equivariant layers). The hierarchical quotient \(\mathcal M_0\supset\cdots\supset\mathcal M_L\) formalizes compositionality and explains the favorable sample complexity of deep, symmetry‑respecting models~\citep{poggioWhyWhenCan2016,poggioCompositionalSparsityLearnable2024}, while clarifying why a manifold prior alone can be insufficient~\citep{kiani2024hardness}. We provide a fuller description of the formalization of hierarchy in Appendix~\ref{ssec:what_hierarchy_means}, as well as a conceptual example in Appendix~\ref{ssec:blender_mapping}. We provided a tentative implementation of predictive coding using this formalism, where the hierarchical structure of the generative model is used ``backwards" (see Appendix~\ref{ssec:lie_hierarchy}). Finally, although we did not discuss this here, methods for symmetry discovery~\citep{Moskalev2022-qf,huSymmetryDiscoveryDifferent2024} provide a route to \emph{learn} \(G\) from data, closing the loop between structure and learning.

\section{Conclusions and Future Directions}
\label{sec:conclusion}

By defining generative models using group theory, we linked compositional symmetry, Lie pseudogroups, and hierarchical reduction to a precise dynamical picture of world‑tracking.  Although this is certainly possible in some cases (e.g., tracking robotic, jointed cats), it may fail when the generative model's latent space is very complex. 
Future work may include: (i) generalization to stochastic inputs and analysis of robustness (SDE analogues of \eqref{eq:feedback}); (ii) development of $K$ operators for valid Lyapunov world tracking problems; (iii) empirical tests with equivariant architectures under controlled generative symmetries; (iv) formal links to Spencer exactness, moduli stacks and integrability guarantees in practical learning systems.

\acks{This work was partly funded by the European Commission under the European Union’s Horizon 2020 research and innovation programme Grant Number 101017716 (Neurotwin) and European Research Council (ERC Synergy Galvani) under the European Union’s Horizon 2020 research and innovation program Grant Number~855109. } 
\bibliography{references}

\begin{thebibliography}{30}
\providecommand{\natexlab}[1]{#1}
\providecommand{\url}[1]{\texttt{#1}}
\expandafter\ifx\csname urlstyle\endcsname\relax
  \providecommand{\doi}[1]{doi: #1}\else
  \providecommand{\doi}{doi: \begingroup \urlstyle{rm}\Url}\fi

\bibitem[Anselmi and Poggio(2022)]{anselmiRepresentationLearningSensory2022}
Fabio Anselmi and Tomaso Poggio.
\newblock Representation learning in sensory cortex: {A} theory.
\newblock \emph{IEEE access : practical innovations, open solutions},
  10:\penalty0 102475--102491, 2022.
\newblock Publisher: Institute of Electrical and Electronics Engineers (IEEE).

\bibitem[Biggs(1974)]{biggs74}
Norman~L. Biggs.
\newblock \emph{Algebraic graph theory}.
\newblock Cambridge University Press, Cambridge, 1974.
\newblock ISBN 0-521-20335-X.

\bibitem[Cagnetta et~al.(2024)Cagnetta, Petrini, Tomasini, Favero, and
  Wyart]{cagnettaHowDeepNeural2024}
Francesco Cagnetta, Leonardo Petrini, Umberto~M. Tomasini, Alessandro Favero,
  and Matthieu Wyart.
\newblock How {Deep} {Neural} {Networks} {Learn} {Compositional} {Data}: {The}
  {Random} {Hierarchy} {Model}.
\newblock \emph{Physical Review X}, 14\penalty0 (3):\penalty0 031001, July
  2024.
\newblock \doi{10.1103/PhysRevX.14.031001}.
\newblock URL \url{https://link.aps.org/doi/10.1103/PhysRevX.14.031001}.
\newblock Publisher: American Physical Society.

\bibitem[Community(2018)]{blender}
Blender~Online Community.
\newblock Blender - a {3D} modelling and rendering package.
\newblock manual, Blender Foundation, Stichting Blender Foundation, Amsterdam,
  2018.
\newblock URL \url{http://www.blender.org}.

\bibitem[Cover and Thomas(2006)]{coverElementsInformationTheory2006}
Thomas~M. Cover and Joy~A. Thomas.
\newblock \emph{Elements of information theory}.
\newblock John Wiley \& sons, 2 edition, 2006.
\newblock tex.date-added: 2016-10-20 21:00:29 +0000 tex.date-modified:
  2016-10-20 21:00:58 +0000.

\bibitem[Friston(2018)]{fristonDoesPredictiveCoding2018}
Karl Friston.
\newblock Does predictive coding have a future?
\newblock \emph{Nature Neuroscience}, 21\penalty0 (8):\penalty0 1019--1021,
  August 2018.
\newblock ISSN 1097-6256, 1546-1726.
\newblock \doi{10.1038/s41593-018-0200-7}.
\newblock URL \url{https://www.nature.com/articles/s41593-018-0200-7}.

\bibitem[Goldschmidt(1967)]{goldschmidt67}
Hubert Goldschmidt.
\newblock Integrability criteria for systems of nonlinear partial differential
  equations.
\newblock \emph{Journal of Differential Geometry}, 1\penalty0 (3):\penalty0
  269--307, 1967.
\newblock \doi{10.4310/jdg/1214428094}.

\bibitem[Hall(2015)]{hall2015}
Brian~C. Hall.
\newblock \emph{Lie groups, lie algebras, and representations: {An} elementary
  introduction}.
\newblock Springer, 2015.

\bibitem[Hu et~al.(2024)Hu, Li, and Lin]{huSymmetryDiscoveryDifferent2024}
Lexiang Hu, Yikang Li, and Zhouchen Lin.
\newblock Symmetry {Discovery} for {Different} {Data} {Types}, October 2024.
\newblock URL \url{http://arxiv.org/abs/2410.09841}.
\newblock arXiv:2410.09841 [cs].

\bibitem[Humphreys(1972)]{humphreys72}
James~E. Humphreys.
\newblock \emph{Introduction to lie algebras and representation theory},
  volume~9 of \emph{Graduate texts in mathematics}.
\newblock Springer-Verlag, New York, 1972.
\newblock ISBN 3-540-90053-5.

\bibitem[Hydon(2000)]{Hydon2000-kp}
Peter~E Hydon.
\newblock \emph{Cambridge texts in applied mathematics: {Symmetry} methods for
  differential equations: {A} beginner's guide series number 22}.
\newblock Cambridge texts in applied mathematics. Cambridge University Press,
  Cambridge, England, January 2000.

\bibitem[Kiani et~al.(2024)Kiani, Wang, and Weber]{kiani2024hardness}
Bobak~T. Kiani, Jason Wang, and Melanie Weber.
\newblock Hardness of learning neural networks under the manifold hypothesis,
  2024.
\newblock URL \url{https://arxiv.org/abs/2406.01461}.
\newblock arXiv: 2406.01461 [cs.LG].

\bibitem[Knapp(2002)]{knapp02}
Anthony~W. Knapp.
\newblock \emph{Lie groups beyond an introduction}, volume 140 of
  \emph{Progress in mathematics}.
\newblock Birkhäuser, Boston, 2 edition, 2002.
\newblock ISBN 0-8176-4259-5.

\bibitem[Li and Vitanyi(2007)]{liApplicationsAlgorithmicInformation2007}
Ming Li and Paul~M.B. Vitanyi.
\newblock Applications of algorithmic information theory.
\newblock \emph{Scholarpedia}, 2\penalty0 (5):\penalty0 2658, 2007.
\newblock tex.date-added: 2016-10-15 14:12:39 +0000 tex.date-modified:
  2016-10-22 13:13:18 +0000.

\bibitem[Lynch and Park(2017)]{Lynch2017-jz}
Kevin~M Lynch and Frank~C Park.
\newblock \emph{Modern robotics}.
\newblock Cambridge University Press, Cambridge, England, May 2017.

\bibitem[Miao and Rao(2007)]{miaoLearningLieGroups2007}
Xu~Miao and Rajesh P~N Rao.
\newblock Learning the {Lie} groups of visual invariance.
\newblock \emph{Neural Computation}, 19\penalty0 (10):\penalty0 2665--2693,
  October 2007.
\newblock Publisher: MIT Press - Journals.

\bibitem[Moskalev et~al.(2022)Moskalev, Sepliarskaia, Sosnovik, and
  Smeulders]{Moskalev2022-qf}
Artem Moskalev, Anna Sepliarskaia, Ivan Sosnovik, and Arnold Smeulders.
\newblock {LieGG}: {Studying} learned {Lie} group generators.
\newblock \emph{arXiv}, 2022.
\newblock Publisher: arXiv.

\bibitem[Poggio et~al.(2016)Poggio, Mhaskar, Rosasco, Miranda, and
  Liao]{poggioWhyWhenCan2016}
T~Poggio, Hrushikesh Mhaskar, Lorenzo Rosasco, Brando Miranda, and Qianli Liao.
\newblock Why and when can deep – but not shallow –{Networks} avoid the
  curse of dimensionality: a review.
\newblock \emph{CBMM Memo}, \penalty0 (058), 2016.
\newblock tex.date-added: 2016-12-04 00:19:36 +0000 tex.date-modified:
  2016-12-04 00:21:03 +0000.

\bibitem[Poggio and Fraser(2024)]{poggioCompositionalSparsityLearnable2024}
Tomaso Poggio and Maia Fraser.
\newblock Compositional sparsity of learnable functions.
\newblock \emph{Bulletin of the American Mathematical Society}, 61\penalty0
  (3):\penalty0 438--456, July 2024.
\newblock ISSN 0273-0979, 1088-9485.
\newblock \doi{10.1090/bull/1820}.
\newblock URL
  \url{https://www.ams.org/bull/2024-61-03/S0273-0979-2024-01820-5/}.

\bibitem[Poggio et~al.(2020)Poggio, Banburski, and
  Liao]{poggioTheoreticalIssuesDeep2020}
Tomaso Poggio, Andrzej Banburski, and Qianli Liao.
\newblock Theoretical issues in deep networks.
\newblock \emph{Proceedings of the National Academy of Sciences}, 117\penalty0
  (48):\penalty0 30039--30045, December 2020.
\newblock \doi{10.1073/pnas.1907369117}.
\newblock URL \url{https://www.pnas.org/doi/10.1073/pnas.1907369117}.
\newblock Publisher: Proceedings of the National Academy of Sciences.

\bibitem[Riesenhuber and Poggio(1999)]{riesenhuberHierarchicalModelsObject1999}
M.~Riesenhuber and T.~Poggio.
\newblock Hierarchical models of object recognition in cortex.
\newblock \emph{Nature neuroscience}, 2:\penalty0 1019--1025, 1999.
\newblock tex.date-added: 2016-05-06 20:19:53 +0000 tex.date-modified:
  2016-05-06 20:23:30 +0000.

\bibitem[Ruffini(2016)]{ruffiniModelsNetworksAlgorithmic2016}
Giulio Ruffini.
\newblock Models, networks and algorithmic complexity.
\newblock \emph{arxiv}, 2016.
\newblock Publisher: arXiv.

\bibitem[Ruffini(2017)]{ruffiniAlgorithmicInformationTheory2017}
Giulio Ruffini.
\newblock An algorithmic information theory of consciousness.
\newblock \emph{Neuroscience of Consciousness}, 2017\penalty0 (1):\penalty0
  nix019, 2017.
\newblock ISSN 2057-2107.
\newblock \doi{10.1093/nc/nix019}.

\bibitem[Ruffini and Lopez-Sola(2022)]{ruffiniAITFoundationsStructured2022}
Giulio Ruffini and Edmundo Lopez-Sola.
\newblock {AIT} foundations of structured experience.
\newblock \emph{Journal of Artificial Intelligence and Consciousness},
  9\penalty0 (2):\penalty0 153--191, September 2022.
\newblock Publisher: World Scientific Pub Co Pte Ltd.

\bibitem[Ruffini et~al.(2024)Ruffini, Castaldo, Lopez-Sola, Sanchez-Todo, and
  Vohryzek]{ruffiniAlgorithmicAgentPerspective2024}
Giulio Ruffini, Francesca Castaldo, Edmundo Lopez-Sola, Roser Sanchez-Todo, and
  Jakub Vohryzek.
\newblock The {Algorithmic} {Agent} {Perspective} and {Computational}
  {Neuropsychiatry}: {From} {Etiology} to {Advanced} {Therapy} in {Major}
  {Depressive} {Disorder}.
\newblock \emph{Entropy}, 26\penalty0 (11):\penalty0 953, November 2024.
\newblock ISSN 1099-4300.
\newblock \doi{10.3390/e26110953}.
\newblock URL \url{https://www.mdpi.com/1099-4300/26/11/953}.
\newblock Number: 11 Publisher: Multidisciplinary Digital Publishing Institute.

\bibitem[Ruffini et~al.(2025)Ruffini, Castaldo, and
  Vohryzek]{ruffiniStructuredDynamicsAlgorithmic2025}
Giulio Ruffini, Francesca Castaldo, and Jakub Vohryzek.
\newblock Structured {Dynamics} in the {Algorithmic} {Agent}.
\newblock \emph{Entropy}, 27\penalty0 (1):\penalty0 90, January 2025.
\newblock ISSN 1099-4300.
\newblock \doi{10.3390/e27010090}.
\newblock URL \url{https://www.mdpi.com/1099-4300/27/1/90}.
\newblock Number: 1 Publisher: Multidisciplinary Digital Publishing Institute.

\bibitem[Seiler(2010)]{seiler10}
Werner~M. Seiler.
\newblock \emph{Involution: {The} formal theory of differential equations and
  its applications in computer algebra}, volume~24 of \emph{Algorithms and
  computation in mathematics}.
\newblock Springer Berlin Heidelberg, Berlin, 2010.
\newblock ISBN 978-3-642-01286-0.
\newblock \doi{10.1007/978-3-642-01287-7}.

\bibitem[Simon(1991)]{simonArchitectureComplexity1991}
Herbert~A. Simon.
\newblock The {Architecture} of {Complexity}.
\newblock In George~J. Klir, editor, \emph{Facets of {Systems} {Science}},
  pages 457--476. Springer US, Boston, MA, 1991.
\newblock ISBN 978-1-4899-0718-9.
\newblock \doi{10.1007/978-1-4899-0718-9_31}.
\newblock URL \url{https://doi.org/10.1007/978-1-4899-0718-9_31}.

\bibitem[Sober(2015)]{soberOckhamsRazorsUsers2015}
Elliott Sober.
\newblock \emph{Ockham's razors: a user's manual}.
\newblock Cambridge University Press, Cambridge, 2015.
\newblock ISBN 978-1-107-06849-0 978-1-107-69253-4.

\bibitem[Sontag(2008)]{sontagInputStateStability2008}
Eduardo~D. Sontag.
\newblock Input to {State} {Stability}: {Basic} {Concepts} and {Results}.
\newblock In Andrei~A. Agrachev, A.~Stephen Morse, Eduardo~D. Sontag,
  Héctor~J. Sussmann, Vadim~I. Utkin, Paolo Nistri, and Gianna Stefani,
  editors, \emph{Nonlinear and {Optimal} {Control} {Theory}: {Lectures} given
  at the {C}.{I}.{M}.{E}. {Summer} {School} held in {Cetraro}, {Italy} {June}
  19–29, 2004}, pages 163--220. Springer, Berlin, Heidelberg, 2008.
\newblock ISBN 978-3-540-77653-6.
\newblock \doi{10.1007/978-3-540-77653-6_3}.
\newblock URL \url{https://doi.org/10.1007/978-3-540-77653-6_3}.

\end{thebibliography}

\appendix \clearpage

\section{Hiearchy}
\label{ssec:what_hierarchy_means}

In our setting, \emph{hierarchy} is the structured way to
decompose a (local) symmetry action and its induced state geometry.

\paragraph{Algebraic factorization (compositional generators).}
Locally (near the identity) we write each world transformation as an ordered product
\begin{equation}
\label{eq:hier_factorization}
\gamma \;=\; \gamma^{(L)} \gamma^{(L-1)} \cdots \gamma^{(1)},
\qquad
\gamma^{(k)} \in G_k ,
\end{equation}
where each $G_k$ is a (finite‑parameter) Lie sub‑pseudogroup generated by a small set of
``level‑$k$'' infinitesimal symmetries.  This is the formal expression of compositionality:
complex deformations are built by nesting a few primitive moves close to the identity.
A concrete instance is the product‑of‑exponentials model in robot kinematics, where each
joint contributes a one‑parameter subgroup and the chain is an iterated semidirect product
\citep{Lynch2017-jz}.

\paragraph{Flag of sub‑pseudogroups (omitting generators).}
Choosing the order in \eqref{eq:hier_factorization} induces a flag
\begin{equation}
\label{eq:hier_flag}
G \;=\; H_0 \;\supset\; H_1 \;\supset\; \cdots \;\supset\; H_L,
\qquad
H_k \;:=\; G_L G_{L-1}\cdots G_{k+1},
\end{equation}
obtained by successively \emph{omitting} generators.  Each step $H_{k-1}\!\to H_k$
declares a new set of symmetry directions ``already explained'' at coarser scale.  The
quotients $H_{k-1}/H_k$ collect the \emph{residual} directions that remain to be explained
at level $k$.

\paragraph{Geometric picture: orbit leaves and nested quotients.}
The action of $H_k$ on the agent’s state space defines a foliation by $H_k$‑orbits.  Passing
to the orbit space produces a nested sequence of \emph{reduced manifolds}
\begin{equation}
\label{eq:hier_quotients}
\mathcal M_0 \;\supset\; \mathcal M_1 := \mathcal M_0/H_1 \;\supset\; \cdots \;\supset\;
\mathcal M_L := \mathcal M_{L-1}/H_L ,
\end{equation}
which we interpret as the agent’s \emph{hierarchical state representation}.  Under static
inputs, world‑tracking fixes the Noether‑style labels associated with the omitted generators,
so trajectories remain on the corresponding orbit leaf; Eq.~\eqref{eq:hier_quotients}
formalizes our ``nested invariant manifolds.''  The Spencer complex provides the
compatibility conditions ensuring that these layered constraints are integrable (no hidden
obstructions when one moves across levels) \citep{goldschmidt67,seiler10}.

\paragraph{Appearance/semantic parameters are part of the symmetry.}
In our generative view on ``cat", \emph{all} controllable attributes—pose and articulation, shape, illumination,
texture, reflectance, eye/fur color, etc.—live in the configuration space $\mathcal C$ and
are acted upon by the same Lie pseudogroup $G$.  Equivalently, we may (locally) decompose
\[
\mathcal C \;\simeq\; \mathcal C_{\text{pose/artic}} \times \mathcal C_{\text{shape}} \times \mathcal C_{\text{appearance}},
\qquad
G \;\simeq\; G_{\text{pose}} \ltimes G_{\text{shape}} \ltimes G_{\text{appearance}},
\]
and require the generative map to be equivariant with respect to the full action,
\begin{equation}
\label{eq:full_equivariance}
f(\gamma\!\cdot c) \;=\; \rho(\gamma)\!\cdot f(c), \qquad \gamma\in G.
\end{equation}
Here $\rho$ is the induced (generally nonlinear) action on image space, which covers not only
geometric moves but also appearance changes (color/texture fields, reflectance, etc.).
Thus, the same hierarchical flag \eqref{eq:hier_flag} may interleave pose-, shape-, and
appearance‑level generators; when a subgroup in the flag is ``frozen,'' its labels are fixed
and the system descends to the corresponding orbit quotient.  For purely discriminative tasks
(e.g., ``cat/not‑cat''), one may \emph{choose} to quotient out some appearance factors at the
top (yielding invariance).  For world‑tracking, the agent typically remains \emph{equivariant}
to all of them, so that $p(x)\approx I_{\theta(t)}$ follows those changes.

When inputs vary slowly or the internal hypothesis updates, level $k$ produces a small
\emph{residual transformation} in the quotient,
$
\varepsilon_k \;\in\; H_{k-1}/H_k,
$ with $
\varepsilon_k \;=\; \exp(\eta_k^a S_k^a) ,
$
where $\{S_k^a\}$ span the Lie directions present in $H_{k-1}$ but absent from $H_k$, and
$\eta_k^a$ are the \emph{error coordinates}.  These residuals induce drift/updates
\emph{along} $\mathcal M_k$.

The hierarchical construction above does not require coarse‑graining.  In practice, however,
after canonicalizing the incoming signal by the current prediction, it is often useful to apply
a task‑dependent coarse‑graining operator   before passing the
residual upward, to improve noise robustness and ensure that each level sees the information
relevant at its scale.  This ``irons out'' small mismatches without altering the orbit‑quotient
logic.

\section{Generative vs.\ Predictive Hierarchies }
\label{appsec:gen_vs_pred_hierarchy_short}

In the Lie–pseudogroup formulation, hierarchy has two complementary directions:
a \emph{generative} (synthesis) order that goes from coarse to fine, and a
\emph{predictive} (inference) loop in which predictions flow top–down while
errors flow bottom–up.

Near the identity, we factor world transformations as
\begin{equation}
\gamma \;=\; \gamma^{(L)} \gamma^{(L-1)} \cdots \gamma^{(1)},
\qquad \gamma^{(k)}\in A_k,
\end{equation}
with a decreasing flag of cumulative remainders
\begin{equation}
H_k \;:=\; A_{k+1}\cdots A_L,
\qquad H_0=G,\ \ H_L=\{e\},
\end{equation}
so the new directions at level $k$ are (locally) $A_k \simeq H_{k-1}/H_k$.
Quotienting by $H_k$ yields the nested reduced manifolds
\begin{equation}
\mathcal M_0 \;\supset\; \mathcal M_1:=\mathcal M_0/H_1 \;\supset\; \cdots \;\supset\;
\mathcal M_L:=\mathcal M_{L-1}/H_L.
\end{equation}

\paragraph{Generative (synthesis) order: coarse $\to$ fine.}
To \emph{generate} an instance from $I_0$, first fix the coarse orbit/quotient
labels (which leaf of $\mathcal M_1,\mathcal M_2,\dots$), then compose the finer
coset moves in order:
\begin{equation}
I \;=\; \gamma\!\cdot\! I_0
\;=\; \big(\gamma^{(L)}\cdots\gamma^{(1)}\big)\!\cdot\! I_0,
\qquad \gamma^{(k)}\in A_k \simeq H_{k-1}/H_k.
\end{equation}

\paragraph{Predictive (inference) loop: predictions down, errors up.}
At time $t$ level $k$ carries $\hat\gamma_k(t)\in H_k$ and issues a top–down prediction
$\hat I_k=\hat\gamma_k\!\cdot I_0$.
The mismatch with the datum $I_{\theta(t)}$ is explained (for small errors) by a
bottom–up residual in the quotient,
\begin{equation}
\varepsilon_k \;\in\; H_{k-1}/H_k \simeq A_k,
\qquad \varepsilon_k=\exp(\eta_k^a S^{(k)}_a),
\end{equation}
estimated by linearising the action at $\hat I_k$.
Level $k$ updates with the part it can realise inside $H_k$ and passes only the unresolved
quotient directions upward (e.g.\ $\eta_k=P_k\eta_k+Q_k\eta_k$, send $Q_k\eta_k$).

\paragraph{Static vs.\ adiabatic.}
With static inputs, equivariance plus tracking fix the Noether‑style labels of the omitted
generators, so trajectories remain on a leaf $\mathcal M_k$. Under slow drift, these labels
are adiabatic invariants and residuals induce controlled motion \emph{within} $\mathcal M_k$.

\section{From Lie Hierarchy to a Blender Rig (Flow‑style Cat)}
\label{ssec:blender_mapping}

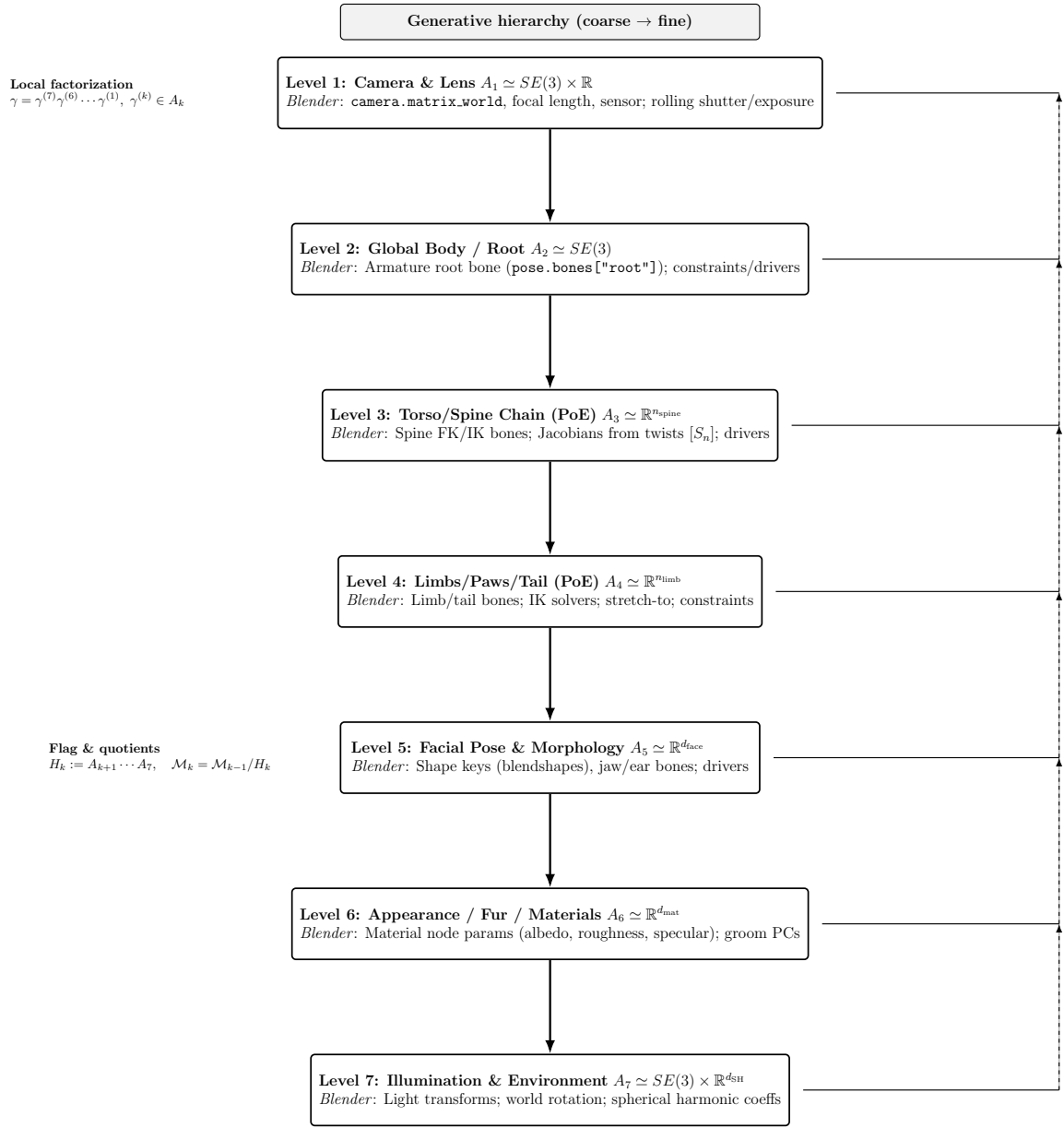
\begin{figure}[t]
\centering
\resizebox{\textwidth}{!}{%
\begin{tikzpicture}[
  x=1cm,y=1cm, transform shape,
  font=\large,
  >=Latex,
  levelbox/.style={
    rounded corners=3pt, draw=black, very thick, align=left,
    fill=white, inner sep=6pt, minimum width=11.2cm, minimum height=1.9cm
  },
  colhead/.style={
    rounded corners=3pt, draw=black, thick, align=center,
    fill=black!5, inner sep=5pt, minimum width=11.2cm, minimum height=0.95cm
  },
  arrowdown/.style={-Latex, ultra thick},
  rail/.style={line width=1pt, draw=black!30},
  errseg/.style={-Latex, thick, dashed},
  hitch/.style={draw=black, line width=0.8pt},
  note/.style={font=\normalsize, align=left}
]

\def\xL{0}        
\def\xE{13.6}     
\def\xLegend{17}  
\def\yTop{0}
\def\dy{2.5}      

\node[colhead] at (\xL,\yTop) {\textbf{Generative hierarchy (coarse $\rightarrow$ fine)}};


\coordinate (start) at (\xL,\yTop-1.9);

\node[levelbox] (L1) at (start) {%
\textbf{Level 1: Camera \& Lens} \hfill $A_1 \simeq SE(3)\times \mathbb{R}$\\
\emph{Blender}: \texttt{camera.matrix\_world}, focal length, sensor; rolling shutter/exposure};

\node[levelbox, below=\dy of L1] (L2) {%
\textbf{Level 2: Global Body / Root} \hfill $A_2 \simeq SE(3)$\\
\emph{Blender}: Armature root bone (\texttt{pose.bones["root"]}); constraints/drivers};

\node[levelbox, below=\dy of L2] (L3) {%
\textbf{Level 3: Torso/Spine Chain (PoE)} \hfill $A_3 \simeq \mathbb{R}^{n_{\text{spine}}}$\\
\emph{Blender}: Spine FK/IK bones; Jacobians from twists $[S_n]$; drivers};

\node[levelbox, below=\dy of L3] (L4) {%
\textbf{Level 4: Limbs/Paws/Tail (PoE)} \hfill $A_4 \simeq \mathbb{R}^{n_{\text{limb}}}$\\
\emph{Blender}: Limb/tail bones; IK solvers; stretch-to; constraints};

\node[levelbox, below=\dy of L4] (L5) {%
\textbf{Level 5: Facial Pose \& Morphology} \hfill $A_5 \simeq \mathbb{R}^{d_{\text{face}}}$\\
\emph{Blender}: Shape keys (blendshapes), jaw/ear bones; drivers};

\node[levelbox, below=\dy of L5] (L6) {%
\textbf{Level 6: Appearance / Fur / Materials} \hfill $A_6 \simeq \mathbb{R}^{d_{\text{mat}}}$\\
\emph{Blender}: Material node params (albedo, roughness, specular); groom PCs};

\node[levelbox, below=\dy of L6] (L7) {%
\textbf{Level 7: Illumination \& Environment} \hfill $A_7 \simeq SE(3)\times \mathbb{R}^{d_{\text{SH}}}$\\
\emph{Blender}: Light transforms; world rotation; spherical harmonic coeffs};

\foreach \a/\b in {L1/L2,L2/L3,L3/L4,L4/L5,L5/L6,L6/L7} {
  \draw[arrowdown] (\a.south) -- (\b.north);
}

\path let \p1=(L1.east) in coordinate (rL1) at (\xE,\y1);
\path let \p1=(L2.east) in coordinate (rL2) at (\xE,\y1);
\path let \p1=(L3.east) in coordinate (rL3) at (\xE,\y1);
\path let \p1=(L4.east) in coordinate (rL4) at (\xE,\y1);
\path let \p1=(L5.east) in coordinate (rL5) at (\xE,\y1);
\path let \p1=(L6.east) in coordinate (rL6) at (\xE,\y1);
\path let \p1=(L7.east) in coordinate (rL7) at (\xE,\y1);

\draw[rail] (rL7) -- (rL1);

\foreach \B/\r in {L1/rL1,L2/rL2,L3/rL3,L4/rL4,L5/rL5,L6/rL6,L7/rL7} {
  \draw[hitch] ($(\B.east)+(0.35,0)$) -- (\r);
}

\draw[errseg] (rL7) -- (rL6);
\draw[errseg] (rL6) -- (rL5);
\draw[errseg] (rL5) -- (rL4);
\draw[errseg] (rL4) -- (rL3);
\draw[errseg] (rL3) -- (rL2);
\draw[errseg] (rL2) -- (rL1);

\node[note] at ($(L1.west)+(-4.8,0)$) {%
\small \textbf{Local factorization}\\[-2pt]
\small $\gamma=\gamma^{(7)}\gamma^{(6)}\cdots\gamma^{(1)},\ \gamma^{(k)}\in A_k$};
\node[note] at ($(L5.west)+(-4.8,0)$) {%
\small \textbf{Flag \& quotients}\\[-2pt]
\small $H_k:=A_{k+1}\cdots A_7,\quad \mathcal M_k=\mathcal M_{k-1}/H_k$};

\end{tikzpicture}%
}
\caption{Blender cat example hierarchy as a Lie‑pseudogroup ladder. Solid arrows (left) show the generative order (coarse $\to$ fine). A right‑hand fine $\to$ coarse \emph{error rail} aggregates bottom‑up prediction residuals (dashed arrows) with short connectors from each level. Each level \(A_k\) is a local group/pseudogroup factor; the flag \(H_k:=A_{k+1}\cdots A_7\) induces nested quotients \(\mathcal M_k\).}
\label{fig:blender_cat_hierarchy_lane}
\end{figure}

A production rig in the software \textsc{Blender} \citep{blender} provides a concrete realisation of our
Lie‑pseudogroup hierarchy for a cat character (as in a scene of the Blender-generated movie \emph{Flow}).
Let $\mathcal C$ collect all controllable model parameters (pose, articulation,
shape, groom, materials, lighting, camera), and let
$f:\mathcal C\to\mathbb R^{X}$ render an image (Cycles/Eevee).  We arrange the
local action of $G$ near the identity as a factorisation
$\gamma=\gamma^{(L)}\cdots\gamma^{(1)}$ with $\gamma^{(k)}\in A_k$,
and define the remainder flag
$H_k:=A_{k+1}\cdots A_L$ (\S\ref{ssec:what_hierarchy_means}).
The table below maps levels to standard Blender constructs (see also Figure~\ref{fig:blender_cat_hierarchy_lane}):

\medskip
\begin{center}
\begin{tabular}{@{}l l l@{}}
\toprule
Level $k$ & Generators $A_k$ (local) & Blender construct (examples)\\
\midrule
1 & Camera $SE(3)$ and lens & \texttt{camera.matrix\_world}, focal length, sensor\\
2 & Global body/root $SE(3)$ & armature root bone (\texttt{pose.bones["root"]})\\
3 & Torso/spine chain (PoE) & spine bones, FK/IK; constraints, drivers\\
4 & Limbs/paws/tail (PoE) & limb bones; IK solvers; stretch‑to constraints\\
5 & Facial morphology & shape keys (blendshapes), jaw/ear bones\\
6 & Fur/appearance & material node params (albedo/roughness); groom P.C.s\\
7 & Illumination gauge & light transforms; spherical‐harmonic coeffs\\
8 & Environment/camera jitter & world rotation; rolling shutter, exposure\\
\bottomrule
\end{tabular}
\end{center}
\medskip

\paragraph{Compositional action (PoE).}
Rigid/articulated motion uses the product‑of‑exponentials (PoE)
\citep{Lynch2017-jz}:
\begin{equation}
T(\theta) \;=\; \Big(\!\prod_{n\in\text{chain}} e^{[S_n]\theta_n}\!\Big)\,M,
\end{equation}
with twists $[S_n]\in\mathfrak{se}(3)$ for bones and $M$ the bone’s home pose.
This is exactly the ordered composition $\gamma^{(k)}\in A_k$ at the geometric
levels (2–4).  Appearance and lighting levels act by nonlinear but smooth
pseudogroup transformations on shader/groom/light parameters; the induced
action $\rho(\gamma)$ on image space remains local
(\S\ref{sec:lie_generative}).

\paragraph{Orbit quotients and nested manifolds.}
Freezing levels up to $k$ (i.e., quotienting by $H_k$) collapses all
configurations equivalent under those generators, yielding the reduced leaf
$\mathcal M_k=\mathcal M_{k-1}/H_k$.  For a static shot, world‑tracking pins
the Noether labels of the omitted generators, so the render/feature readouts
stay constant on $\mathcal M_k$ (cf.\ Eq.~\eqref{eq:hier_quotients}).

\paragraph{Predictive residuals in practice.}
At frame $t$ level $k$ predicts $\hat I_k=\hat\gamma_k\!\cdot I_0$
(rendered from $\hat c_k$).  The error $\delta I_k=I_{\theta(t)}-\hat I_k$ is
explained by a small residual $\varepsilon_k\in H_{k-1}/H_k\simeq A_k$:
linearise the image change along the level‑$k$ generators,
\begin{equation}
\delta I_k \;\approx\; \sum_{a}\eta_k^a\,V^{(k)}_a(\hat I_k),\qquad
\varepsilon_k\simeq \exp(\eta_k^a S^{(k)}_a),
\end{equation}
where $V^{(k)}_a$ are image‑space velocities induced by basis
$S^{(k)}_a\in\mathrm{Lie}(A_k)$ (estimated by finite‐difference renders).
Update $\hat\gamma_k\leftarrow \hat\gamma_k\,\exp(P_k\eta_k)$ using the part
$P_k\eta_k$ realisable within $H_k$, and pass the unresolved component
$Q_k\eta_k$ upward (\S\ref{ssec:predictive_lie}).

\paragraph{Blender blueprint (schematic).}
\emph{Parameterisation.} A minimal state vector can include:
camera (6\,+\,lens), root (6), $\sim$20–40 joint angles (PoE), 10–30
shape‑key coeffs, 3–8 groom principal components, 9 SH lighting coeffs; the
exact split defines the sets $A_k$.
\emph{Application.} Within \texttt{bpy}, each $\gamma^{(k)}$ is applied by
writing pose bone transforms, shape key values, material node sockets, and
light/world transforms, then rendering to a buffer:
\begin{equation*}
\hat I_k \;=\; f\bigl(\gamma^{(k)}\!\cdot\hat c_{k-1}\bigr).
\end{equation*}

\paragraph{Spencer/compatibility as rig constraints.}
Blender’s dependency graph (constraints, drivers, IK) enforces integrability of
the layered actions—precisely the ``no hidden obstruction'' condition captured
abstractly by the Spencer complex \citep{seiler10}.  In practice: no cycles in
drivers; consistent bone/rest matrices; shader/groom parameters driven only by
lower levels or constants.

\paragraph{Why this matches the movie‑making intuition.}
The director’s controls are inherently hierarchical: camera and blocking first,
then body motion, then paws/tail, then facial nuance, then appearance/lighting.
Our $G=H_0\supset\cdots\supset H_L$ flag formalises that workflow: each stage
\emph{defines} a quotient leaf $\mathcal M_k$ (solid arrows in
Fig.~\ref{fig:lie_pipeline_fullpage}), while residuals
$H_{k-1}/H_k$ \emph{move} the solution along that leaf (dashed arrows), until
the frame’s constraints are met (static case) or track a slowly varying target
(adiabatic case).  The PoE levels (2–4) give exact Lie‑group composition;
appearance/lighting act as smooth pseudogroup charts.


\end{document}